\documentclass{article}
\usepackage{graphicx}
\usepackage{amsmath}
\DeclareMathOperator*{\argmax}{arg\,max}
\usepackage{authblk}
\begin{document}

\title{Framewise approach in multimodal emotion recognition in OMG challenge}
\author[1, 2]{\small Grigoriy Sterling}
\author[1, 3]{\small Andrey Belyaev}
\author[1]{\small Maxim Ryabov}

\affil[1]{\footnotesize Neurodata Lab LLC, USA}
\affil[2]{\footnotesize Institute for Information Transmission Problems, Moscow, Russian Federation}
\affil[3]{\footnotesize Moscow State Univercity, Moscow, Russian Federation}

\maketitle

\begin{abstract}
In this report we described our approach achieves $53\%$ of unweighted accuracy over $7$ emotions and $0.05$ and $0.09$ mean squared errors for arousal and valence in OMG emotion recognition challenge. Our results were obtained with ensemble of single modality models trained on voice and face data from video separately. We consider each stream as a sequence of frames. Next we estimated features from frames and handle it with recurrent neural network. As audio frame we mean short $0.4$ second spectrogram interval. For features estimation for face pictures we used own ResNet neural network pretrained on AffectNet database. Each short spectrogram was considered as a picture and processed by convolutional network too. As a base audio model we used ResNet pretrained in speaker recognition task. Predictions from both modalities were fused on decision level and improve single-channel approaches by a few percent.
\end{abstract}

\section{Introduction}
OMG challenge suggests to classify a given video fragments into one of $7$ discrete emotion categories and recognize $2$ dimensional emotion parameters: arousal and valence. The OMG Emotion Behavior dataset description can be found in \cite{omg}. It consist of $178$ long videos devided into $2725$ short utterances of vary length from $0.5$ to $30.6$ seconds. It had been already devided into train, validation and test sets. 

This dataset is not balanced: the least frequent emotion label is surpise with the only $26$ examples in the train set. Therefore to get predictions in the suggested dataset we decided to fine-tune some base models pretrained on other datasets. But before it we need to preprocess data in the right way.

\section{Data preprocessing}
We were focused on the most informationful modalities in emotion recognition problem: face, voice (exclude text information) and body motion. 

Firstly we used Caffe implementation of Faster RCNN \cite{frcnn} to cut face boxes from frames in all videos. After resising to $112$x$112$ pixels we aligned it by lips and nose points, and obscure background. 

A bit more interesting preprocessing we did with audio channel. Our goal was to represent audio stream as a sequence of frames, like video. From raw wav signal we estimated spectrogram with $20$ ms window, $10$ ms overlap and $4000$ Hz upper frequency bound. Next we divided spectrograms into $50\%$ overlapping frames of length $0.4$s. The final frame $x$ is a concatenation of $3$ matrices considered as RGB channels: $[log(x), norm(log(x)), equalized\_hist(log(x))]$, where $log$ is a natural logarithm, $norm(x) = \frac{x - \min x}{\max x - \min x}$ is normalization function and histogram equalization is a contrast transofrmation to make pixel intensities uniform in a given picture. Thus each audio fragment was represented as a sequence of $3$x$40$x$40$ pictures with frequency equals to $5$. 

Examples of face and audio frames after preprocessing are shown below:

\begin{figure}[h!]
\includegraphics[scale=0.4]{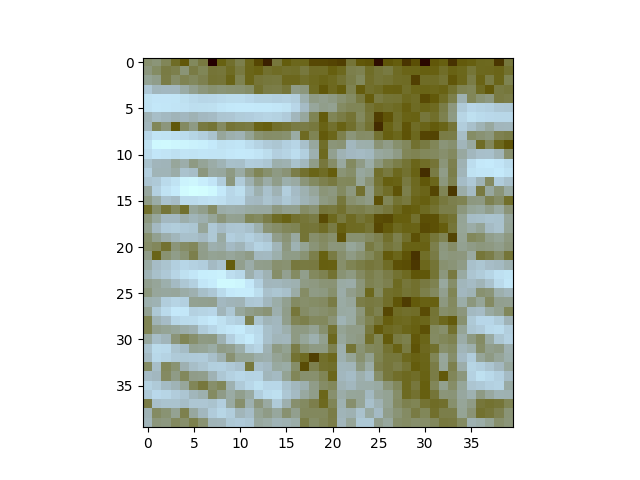}
\includegraphics[scale=0.4]{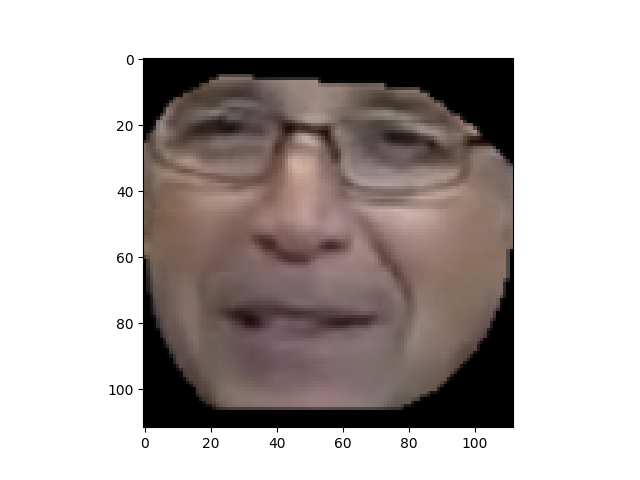}
\end{figure}

In addition to voice and face information we used body keypoints as the third data source, but it seems to be less informationful.

To make predictions for utterances we divided it into $2$s intervals and trained models to predict emotions on these intervals.

\section{Face model}
Since training set is relatively small, we decided to pretrain model on different emotional dataset. We have chosen AffectNet database \cite{affectnet}. The main advantages of this corpus is that emotional face pictures were collected in the wild and have many examples in each class. We trained ResNet-$50$ neural network and achieved $64\%$ accuracy on $7$ classes on it. Next we used this model as a base shared features estimator in the challenge. Next we handle a sequence of frames in each $2$s interval by a recurrent neural network with long short-term memory cells. So, our face model predicts one class of $7$ emotions for a short video fragment. If we average predictions for the whole utterance we achieve $51\%$ unweighted accuracy over $7$ emotions on the test set. 

We have also tried to train this model with Connectionist Temporal Classification (CTC) loss function \cite{ctc, ctc2}, but that did not increased the final score, instead of audio model.

\section{Connectionist Temporal Classification}

Emotionality does not fill any utterances uniformely. In other words, some parts of emotional speech are more emotional than other. One possible way to handle it is CTC loss function. 

Let $\mathbf{y}$ be the $T$x$N$ matrix of RNN outputs, where $T$ is the number of frames in a sequence and $N$ is classes count: $y_{i, t}$ represents the probability of $t$-th frame has $i$-th label. So, the probability to observe a label sequence $\pi$ for an input feature sequence $\mathbf{x}$ is:

\begin{equation}
P(\pi, x) = \prod_{t=1}^T{y_{\pi_t,t}}
\end{equation}

Let c be label sequence and $\mathcal{B}(c, T)$ is a set of all possible labellings of length $T$, that could be reduced to $c$ after removing blank and repeated symbols. For example, $\mathcal{B}(A, 2) = \{-A, A-, AA\}$, $\mathcal{B}(AB, 3) = \{-AB, AAB, AB-, ABB, A-B\}$. Next we can denote a conditional probability of an input object $x$ to have label $L$:

\begin{equation}
P(L, x) = \sum_{\pi \in \mathcal{B}(L, T)}P(\pi, x)
\end{equation}

And the final step is to maximize $P(L^*, x)$ where $L^*$ is the true labelling of $x$. At prediction stage we have no true label $L^*$, but we can estimate the most probable path $\pi^* = \argmax_\pi P(\pi, x)$, that we will use as prediction. An algorithm how to get that argmax was described in \cite{ctc}.

Keeping in mind our task of emotion recognition, CTC approach becomes more specific. In this case possible emotion labels $L$ could be one of the following pattern: $E$, $-E$, $E-$ and $-E-$, where $E$ is emotion label like anger, happy, etc.

\section{Voice model}

For speech channel we went by the same way as for video. Firstly we need to estimate high-level features from colored pictures represents short spectrograms. We found out that features from speaker verification model are more useful than any other. For a given $0.4$s spectrogram frame this model tries to recognize who is an author of it. We achieved $72\%$ top-$1$ and $91\%$ top-$5$ classification accuracy on $1032$ speakers from Ted Talks data \cite{ted}. After clearing audiotracks of noise we got about $1000$ frames per speaker to train and validate. 

Next we train LSTM RNN on a sequence of high-level features estimated for frames in OMG utterances. After averaging prediction at $2$s intervals level our approach gave $49\%$ accuracy using only audio data. 

\section{Multimodal approach}

The final model uses three data channels as inputs and processed it with the corresponding pretrained models. Sequences of high-level features are fed into recurrent parts of our model to get $2$s interval-wise vector representations. At decision-level we concatenated it in a single vector and process with fully connected layer with softmax activation function. The resulting label is a simple argmax of the averaged over the whole utterance predictions. 

\section{Results and conclusion}

Our approach achieved $53\%$ of unweighted accuracy (the corresponding average F1 score is $0.5$) in emotion classification task and $0.04$ and $0.09$ MSE error for arousal and valence. 

We got these results with deep neural network that estimates features from preprocessed pictures via convolutional layers, and they are fed into recurrent layers to handle its evolution over time. For each modality we trained different models and fused it at decision level.

\end{document}